\documentclass{article}

\usepackage{microtype}
\usepackage{graphicx}
\usepackage{subfigure}
\usepackage{booktabs} %

 \usepackage{booktabs}
\usepackage{longtable}
\usepackage{float}
\usepackage{adjustbox}
\usepackage{multicol}
\usepackage{multirow}

\usepackage{amsmath}
\usepackage{hyperref}
\usepackage[capitalize,noabbrev]{cleveref}

\usepackage[accepted]{icml2024}

\usepackage{amssymb}
\usepackage{mathtools}
\usepackage{amsthm}
\usepackage{multirow} 
\usepackage{multicol}

\usepackage{listings}
\usepackage{xcolor}

\definecolor{codegreen}{rgb}{0,0.6,0}
\definecolor{codegray}{rgb}{0.5,0.5,0.5}
\definecolor{codepurple}{rgb}{0.58,0,0.82}
\definecolor{backcolour}{rgb}{0.95,0.95,0.92}

\lstdefinestyle{mystyle}{
    backgroundcolor=\color{backcolour},   
    commentstyle=\color{codegreen},
    keywordstyle=\color{magenta},
    numberstyle=\tiny\color{codegray},
    stringstyle=\color{codepurple},
    basicstyle=\ttfamily\footnotesize,
    breakatwhitespace=false,         
    breaklines=true,                 
    captionpos=b,                    
    keepspaces=true,                 
    numbers=left,                    
    numbersep=5pt,                  
    showspaces=false,                
    showstringspaces=false,
    showtabs=false,                  
    tabsize=2
}

\theoremstyle{plain}

\theoremstyle{definition}

\theoremstyle{remark}

\usepackage[textsize=tiny]{todonotes}

\icmltitlerunning{Emergency Department Decision Support}

\begin{document}

\twocolumn[
\icmltitle{Emergency Department Decision Support using Clinical Pseudo-notes}



\icmlsetsymbol{equal}{*}

\begin{icmlauthorlist}
\icmlauthor{Simon A. Lee}{yyy}
\icmlauthor{Sujay Jain}{zzz}
\icmlauthor{Alex Chen}{yyy}
\icmlauthor{Kyoka Ono}{ddd,eee}
\icmlauthor{Akos Rudas}{yyy}
\icmlauthor{Jennifer Fang}{aaa,bbb,ccc}
\icmlauthor{Jeffrey N. Chiang}{yyy,www}
\end{icmlauthorlist}

\icmlaffiliation{yyy}{Department of Computational Medicine, University of California, Los Angeles, CA 90095 USA}
\icmlaffiliation{www}{Department of Neurosurgery, University of California, Los Angeles, CA 90095 USA} 
\icmlaffiliation{zzz}{Department of Electrical and Computer Engineering, University of California at Los Angeles, Los Angeles, CA 90095 USA}
\icmlaffiliation{aaa}{LA Health Services, Enterprise Clinical Informatics, Los Angeles, CA}
\icmlaffiliation{bbb}{Harbor-UCLA Medical Center, Department of Emergency Medicine, Torrance, CA}
\icmlaffiliation{ccc}{University of California, Los Angeles, Department of Emergency Medicine, Los Angeles, California}
\icmlaffiliation{ddd}{Department of Statistics and Data Science
       University of California, Los Angeles}
\icmlaffiliation{eee}{Department of Natural Sciences
       International Christian University,
       Mitaka, Tokyo, Japan}

\icmlcorrespondingauthor{Simon A. Lee}{simonlee711@g.ucla.edu}
\icmlcorrespondingauthor{Jeffrey N. Chiang}{njchiang@g.ucla.edu}

\icmlkeywords{Machine Learning, ICML}

\vskip 0.3in
]



\printAffiliationsAndNotice{\icmlEqualContribution} 

\begin{abstract}
In this work, we introduce the Multiple Embedding Model for EHR (MEME), an approach that serializes multimodal EHR tabular data into text using ``pseudo-notes'', mimicking clinical text generation. This conversion not only preserves better representations of categorical data and learns contexts but also enables the effective employment of pretrained foundation models for rich feature representation. To address potential issues with context length, our framework encodes embeddings for each EHR modality separately. We demonstrate the effectiveness of MEME by applying it to several {decision support} tasks within the Emergency Department across multiple hospital systems. Our findings indicate that MEME outperforms {traditional machine learning, EHR-specific foundation models, and general LLMs,} { highlighting its potential as a general and extendible EHR representation strategy.} 
\end{abstract}

\section{Introduction}
In recent years, increased access to Electronic Health Records (EHR) has provided healthcare systems with valuable insights into patients' health histories \citep{idowu2023streams}. This wealth of information makes EHR indispensable for inference tasks, especially as the machine learning community increasingly focuses on healthcare applications. Both traditional and cutting-edge machine learning techniques have been harnessed to aid in specific diagnosis and prognosis tasks \citep{shickel_deep_2018}, with a recent keen interest in incorporating state-of-the-art foundation models, such as Large Language Models (LLMs) \citep{zhao_survey_2023}. These advanced models, pre-trained on extensive and diverse textual corpora, offer a broad understanding across numerous domains, enhancing their adaptability and effectiveness for various complex tasks. Notably, their ability to generate high-fidelity latent representations can be directly employed in many classification tasks, often demonstrating state-of-the-art performance. However, the adoption of these models, built for natural language and text, has been hampered by the fact that the canonical form of generally available EHR data is tabular, and access to textual clinical notes is generally infeasible due to privacy concerns.

We are particularly interested in integrating Electronic Health Records with Large Language Models because traditional machine learning paradigms often struggle with the heterogeneous nature of EHR data. EHRs are disparate data sources, encompassing a wide array of data types, including numerical (e.g., lab test results), categorical (e.g., diagnosis codes, medication types), and free-text data, while spanning multiple biological domains and scales. This diversity enables comprehensive insights into patient health, treatments, and outcomes but also presents numerous challenges. For instance, EHRs frequently include categories with a large number of classes. Traditional methods like one-hot encoding and conventional feature engineering can obscure the inherent meanings within this data, leading to issues such as inefficient data ingestion, a lack of contextual understanding, and sparse data representations. Addressing these challenges is crucial for making full use of the rich, yet complex, information EHR systems contain.

 \begin{figure*}[t]
   \centering 
   \includegraphics[width=6in]{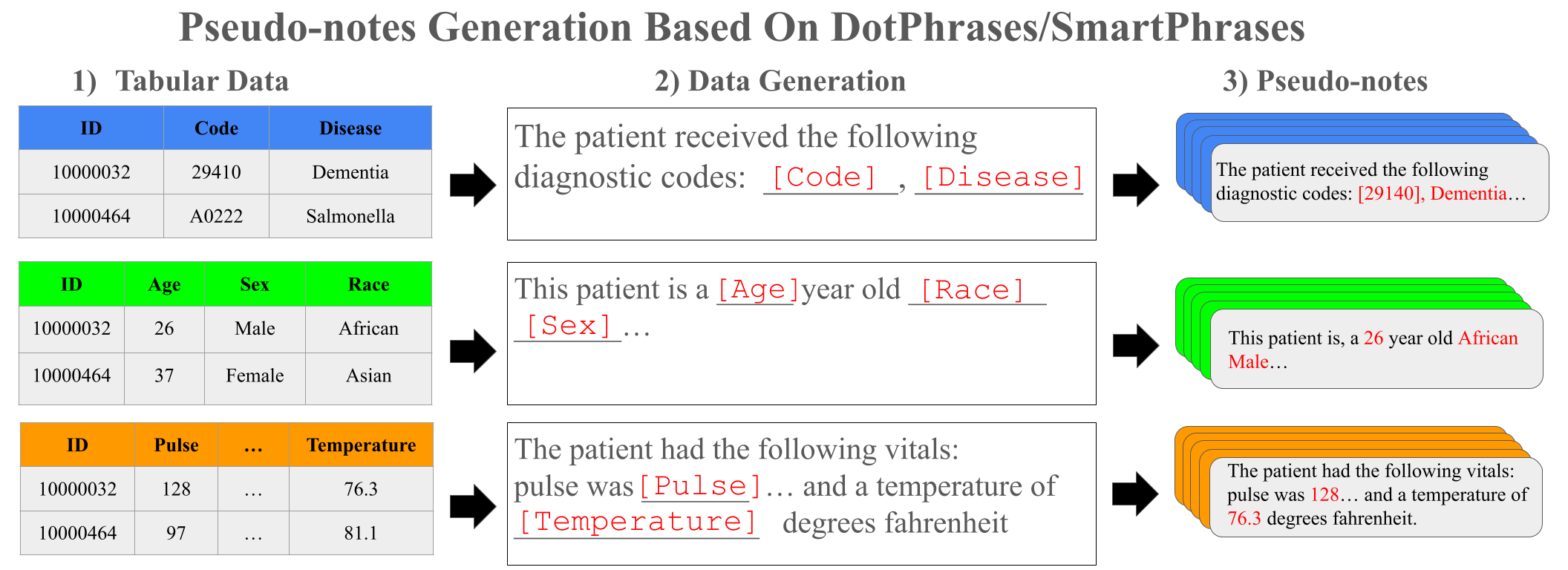} 
   \caption{An Overview of Clinical Pseudo-Notes Generation: From Tabular Data to Text Using DotPhrases/SmartPhrases Commonly Employed in Healthcare.}
   \label{fig11} 
 \end{figure*} 

Therefore, in this work, we introduce the \textit{Multiple Embedding Model for EHR (MEME)}, an approach that processes tabular health records and generates textual representations via our ``pseudonotes'' method. Using these clinical pseudo-notes, MEME bridges the gap between tabular EHR data and modern Natural Language Processing (NLP) techniques. We also find that adopting a multiple embedding strategy, where different components of the EHR are encoded separately, yields better results compared to trying to fit all our text into a single heterogenous embedding. We demonstrate this utility by benchmarking MEME against other EHR methods on prediction tasks related to the emergency department. This is an important problem in both machine learning and healthcare because utilizing a patients health history can help predict future events or requirements needed from the emergency department with high precision.

\subsection*{Generalizable Insights about Machine Learning in the Context of Healthcare}

In this work, we evaluate text serialization as an interface between tabular electronic health records and large language foundation models. We find our multimodal text serialization approach which separately considers EHR components and leverages the general reasoning capacity of language foundation models outperforms existing  models specifically tailored for healthcare, and that task-specific tuning further outperforms prompting-based approaches. While we demonstrate these capabilities on several benchmark tasks in the context of the Emergency Department, we expect these findings to be applicable to other decision support settings. We also observe limitations in the context of cross-site model generalizability.

\section{Related Work}

\subsection{LLM \& Tabular Data}

An ongoing research area involves applying LLMs to tabular data. Several works have focused on using canonical tabular data with existing foundational models, as referenced in \citep{zhang2023towards} and \citep{slack2023tablet}. Additional efforts include utilizing EHR in their canonical form with foundational models, as seen in (\cite{shi2024ehragent}; \cite{wang2023meditab}). A more recent approach involves constructing patient summaries directly from tabular data using LLMs for natural integration into future NLP tasks, highlighted by \citep{ellershaw2024automated} and \citep{hegselmann2024data}. However, such methods have not accounted for the potential for hallucinations by these LLMs, especially in medical applications, as highlighted by \citep{lee2024large}.

Therefore, researchers have explored better methods for converting tabular data into textual formats that harmonize better with LLMs. There are current efforts proposed by \citep{arnrich2024medical} to have some standard for EHR data (Medical Event Data Standard) which comply with existing models. Other previous works \citep{hegselmann2023tabllm}to represent tabular data have led to the development of a novel concept referred to as ``stringified" or serialized tabular data. This technique transforms tabular data into a text-based format, either as a simplified list (e.g., Age: 42, Height: 143cm, ...) or through serialized sentences. Such a transformation allows for a more seamless integration of diverse data types into language models, enabling their analysis with advanced machine learning techniques. The growing interest in this area has facilitated the application of state-of-the-art language models to tabular data, often achieving superior performance compared to traditional machine learning models in scenarios with minimal or no training data. This capability has been demonstrated through a text-to-text prompt-based approach that uses serialized tabular data, leveraging the vast knowledge encapsulated within the parameters of LLMs for various zero-shot and few-shot learning tasks, as illustrated in \citep{hegselmann2023tabllm}. 

Moreover, another study explored the integration of paired datasets (e.g., images) for game data, incorporating both tabular textual and visual fields cohesively for predictive tasks. This was achieved using BERT models to handle the textual component of the datasets \citep{lu2023MuG}, showcasing their versatility in processing complex data structures and blending different types of information seamlessly. Such integrative approaches suggest new possibilities in applying LLMs beyond traditional text-based applications.

\subsection{LLMs, Transformers \& EHR}
Transformer based models and LLMs have also been increasingly applied to EHR data \citep{kalyan_ammu_2022}. These applications tend to fit in one of two paradigms as identified by \citep{wornow_shaky_2023}: One approach that operates upon structured information captured within the EHR and another that operates upon clinical text with recent methods involving multimodal approaches. 

Models developed to represent structured EHR data are generally based on the BERT architecture \citep{kalyan_ammu_2022}. Broadly speaking, these approaches represent the electronic health record as a sequence of events (\cite{wornow2024ehrshot}; \cite{hur2023genhpf}; \cite{hur2022unifying}). For example, BEHRT and its derivatives CEHR-Bert, construct sequences of diagnostic codes and visit metainformation to be compatible with the BERT framework \citep{li_behrt_2020, li_hi-behrt_2021, rasmy_med-bert_2021, pang_cehr-bert_nodate}. They are able to predict future outcomes of visits following a in-context learning/chain-of-thought (CoT) approach, now very popularized in LLM research \citep{wei2022chain}. These models are better adapted to currently available healthcare data due to deidentification and standardization efforts (e.g., OHDSI/OMOP \citep{noauthor_omop_nodate}). 

Notably, nearly all approaches treat EHR data as a single heterogeneous data source, despite the fact that EHR covers data across multiple biological scales and clinical domains (e.g., billing-related diagnostic codes, molecular blood tests, vital-sign measurements). Indeed, more recently, EXBEHRT found that separately representing the components of EHR offers several benefits, including improved performance, shortened sequence length, and fewer required parameters \citep{rupp_exbehrt_2023}.

Another approach directly operates upon clinical notes, which are in the form of natural text. The primary advantage of these approaches is the ability to incorporate pretrained models, such as GPT, BioBert, PubMedBert, etc. \citep{roy-pan-2021-incorporating, chief-bert}. However, these data are challenging to acquire, with nearly all applications restricted to a single database (MIMIC-III, \citep{johnson_mimic-iii_2016}). Recently, efforts have been made to begin multimodal analysis, where groups analyze different data from the EHR (imaging, waveforms, etc.) for predictive tasks. MC-BEC uses all available data within their in-house dataset and performs a multilabel classification on various EHR tasks \citep{chen2023multimodal}. Other attempts have unified imaging data with text \citep{khader2023medical}, and imaging data with structured EHR for various analyses \citep{zhou2023transformer}.

Therefore, in this work, with the continued development of transformer and LLM methods, we develop our own approach that can capture a joint representation of multimodal (e.g. arrival, triage, etc.) structured EHR data. We do so by constructing ``pseudo-notes" out of raw EHR tabular data contained within the MIMIC-IV and our institutional dataset. We then feed this data into a foundation model encoder to obatin representations before finally processing it through a feed forward network with self attention to train a classifier model to predict various binary classification tasks related to the emergency department. 


\section{Benchmarks}
\label{data}

This paper focuses on binary prediction tasks around Emergency Department disposition and decompensation defined in \citep{chen2023multimodal}. We test our multimodal method's ability to be used in both single classification and multilabel classification tasks benchmarked against other tabular-based and textual-operating machine learning models. This assessment aims to demonstrate the performance advantages of adopting a text-based and multiple embedding strategy.
\subsection{Benchmark tasks}

\begin{enumerate}
        \item \textbf{ED Disposition (Binary Classification):} Our first objective is to predict ED Disposition, determining where patients were sent after their Emergency Room visit, based on EHR measurements recorded during their stay in the ED. We frame this as a binary classification problem, distinguishing whether the patient was discharged home or admitted to the hospital.
        
        \item \textbf{ED Decompensation\iffalse Tasks \fi (Multilabel Binary Classification):} Our subsequent objective is to analyze the subset of patients admitted to the hospital and predict various other measures related to the ED. In this set of tasks, we adopt a multilabel binary classification approach, where the model predicts three separate ED tasks simultaneously. The first task involves predicting the patient's next discharge location, distinguishing between home and other facilities (not home). The second task is predicting the requirement for Intensive Care Unit (ICU) admission.
        And the last task
        is predicting patient mortality, specifically whether the patient dies during their 
        hospital stay.  
\end{enumerate}

\begin{table}[h!]
\caption{Prevalence Statistics for Different Benchmark Tasks in the MIMIC-IV and Institutional Datasets, indicating the proportion of samples in which these outcomes occurred.}
\label{prevalence-table}
\begin{center}
\begin{small}
\begin{sc}
\begin{tabular}{lcc}
\toprule
Task & MIMIC-IV & Institutional \\
\midrule
ED Disposition & 0.395 & 0.253 \\
\bottomrule
Discharge Location & 0.449 & 0.381 \\
ICU & 0.197 & 0.157 \\
Mortality & 0.029 & 0.031 \\
\bottomrule
\end{tabular}
\end{sc}
\end{small}
\end{center}
\end{table}

 In Table \ref{prevalence-table}, we detail the classification objectives and provide a summary of dataset statistics related to the prevalence of each label in these tasks.

\subsection{Data Source}

Our study sources data from the Medical Information Mart for Intensive Care (MIMIC)-IV v2.2 database \citep{johnson_mimic-iv_nodate} 
and further evalution is done on our Institutional EHR database. 
We detail the components of this database to further explore the data inputs for our model. 


\begin{figure*}[t]
   \centering 
   \includegraphics[width=6in]{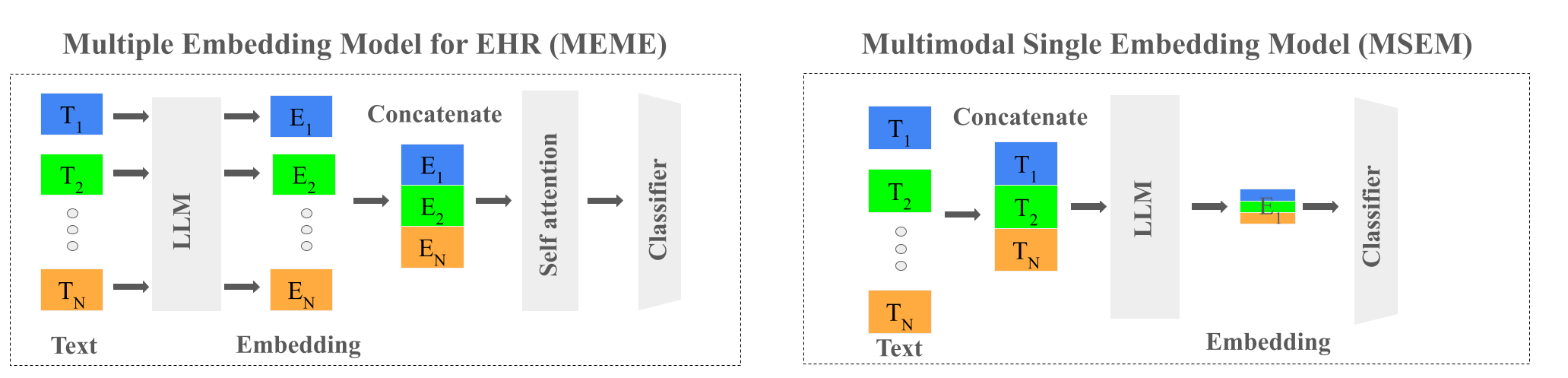} 
   \caption{The Multiple Embedding Model for EHR (MEME) and the Multimodal Single Embedding Model (MSEM) side by side.}
   \label{arc} 

 \end{figure*} 

\begin{itemize}
    \item \textbf{MIMIC-IV ED \citep{Johnson2023MIMICIVED}:} For these downstream tasks, the EHR concepts (modalities) utilized include: \textit{arrival information}, capturing patient demographics and means of arrival; \textit{triage}, documenting patient vitals and complaints at the time of arrival; \textit{medication reconciliation (medrecon)}, detailing prior and current medications taken by the patient; \textit{diagnostic codes} (ICD-9/10 codes) assigned for diagnosis; as well as measurements collected throughout the ED stay, including \textit{patient vitals} and medications received from \textit{pyxis}. All data points across modalities can be combined using a unique Visit or Hospital admission ID (Hadm\_id) and can also be connected to all prediction labels. 
    \item \textbf{Institutional Database:} 
    In our Institutional data, we have access to all data modalities from the MIMIC-IV database with the exception of medication reconciliation (medrecon). However, there may be slight variations in the data across modalities due to the lack of consistency in features in different EHR systems. Our approach is to model our pseudo-notes so that they closely resemble each other. Similar to the MIMIC-IV database, all modalities in our institutional data can be linked using a hospital admission ID and are also associated with all prediction labels.

\end{itemize}

\vspace{-0.2cm}

In the MIMIC-IV database, we analyzed $400,019$ unique visits, each associated with six modalities, contributing to a dataset size of approximately $2.4$ million text paragraphs. For predicting ED Disposition, we use the available data for training, validation, and testing with a set seed for reproducibility purposes. For the three decompensation prediction tasks, we utilize the subset of visits who were admitted to the hospital from the ED, resulting in a sample size of $158,010$ patients. Additionally, in the institutional database, we have a much larger sample size of $947,028$ patients with 5 available modalities (excluding \textit{medrecon}), resulting in approximately $4.75$ million text paragraphs derived from the EHR. We use all available data for the ED disposition task, and the $240,161$ admitted patients were used for decompensation prediction. Further breakdowns can be seen in our strobe diagrams attached in the appendix (Section \ref{strober}).

\section{Methods}

\subsection{Clinical Pseudonotes Method}
\label{notes}

We begin by detailing the pseudo-notes method, which was developed in collaboration with a clinical informaticist at our institution. Although various approaches for text generation using Large Language Models (LLMs) have been explored \citep{hegselmann2024data} \citep{ellershaw2024automated}, we identify potential pitfalls in current methodologies due to issues related to hallucinations, as described in \citep{hegselmann2024data} and \citep{lee2024large}. To address these challenges, we employ a more traditional serialization process mimicking SmartPhrases/DotPhrases on the Epic
EHR system \citep{chang2021emr}, which assists in accurately filling in relevant clinical information within a fixed passage to generate the pseudo-notes as shown in Figure \ref{fig11}. We construct separate notes for each clinical modality (e.g., arrival, triage, etc.), resulting in six paragraphs for each patient visit. Additionally, we omit any patient identifiers within the text to mitigate the risk of overfitting on these unique identifiers (e.g., Patient ID, Hospital Admission ID, etc.). While there may be concerns surrounding the similarity between texts, we find that running a BERTopic \citep{grootendorst2022bertopic}, which is analogous to Latent Dirichlet Allocation (LDA), finds unique topics within our corpus of pseudo-notes text. Example Pseudo-notes are attached in the Appendix (\ref{exnotes}).

\subsection{Multiple Embedding Model for EHR (MEME) Architecture}
In this section, we outline our model architecture, which is crucial to our end-to-end pipeline. These networks are tailored to process preprocessed and tokenized textual inputs, outputting logits for each class. The class with the highest probability is then selected as the predicted class. We begin by explaining that this approach is designed to embed a patient's modalities separately, accommodating the variable textual lengths associated with different patient visits as described in \citep{rupp_exbehrt_2023}. This strategy addresses a limitation of the BERT architecture, which has a sequence limit of 512 tokens. This method is preferable to the truncation involved in fitting all text into a single heterogeneous embedding, as it better preserves the integrity of the input data and it omits any concerns regarding different ordering strategies. A direct comparison of these approaches can be seen in Figure \ref{arc} which shows our proposed architecture (MEME) over the conventional single heterogenous embedding approach (MSEM).

\subsubsection{Step 1: Generating Embeddings}

In the initial step of our model, we aim to generate embeddings for each EHR concept by feeding tokenized data into our foundational models' encoders, which produce rich, high-dimensional vector representations encapsulating various aspects of a patient's medical history. We choose to freeze the encoder layers, focusing on the training parameters of the subsequent layers dedicated to the prediction task. After generating embeddings for all concepts, we concatenate them into a unified input vector for further processing. This procedure can be mathematically represented as follows: In the model's first phase, modality-specific pseudo-notes are processed and structured into a tokenized format, denoted $D_{\text{tokenized}}$, which outlines a series of unique medical concepts or characteristics ($c_i$) derived from a patient's records. Each concept undergoes transformation via the foundation models' encoder into a high-dimensional vector $\vec{v}_i$, capturing clinical information via context-rich portrayals of each EHR concept. These vectors are then unified into a comprehensive vector $\vec{V}_{\text{concat}}$ through concatenation, laying the groundwork for our multimodal patient embeddings.

\begin{equation}
    \vec{v}_i = \text{FoundationModel}(c_i) \quad \forall c_i \in D_{\text{tokenized}}
\end{equation}
\begin{equation}
    \vec{V}_{\text{concat}} = \text{Concatenate}(\vec{v}_1, \vec{v}_2, \ldots, \vec{v}_n)
\end{equation}

\subsubsection{Step 2: Self-attention Classifier}

In the second step of our network, we introduce a new use case of a self-attention layer \citep{vaswani2017attention} designed to analyze the singular concatenated representation vector, $\vec{V}_{\text{concat}}$, as a unified entity. This approach arises from our intention to interpret aligned modalities collectively, rather than as separate entities, allowing the network to operate comprehensively on the entire vector. It evaluates the relationships between elements within the vector, capturing patterns across different EHR concept vectors. 
The output from this layer is then directed through a fully connected layer, followed by a ReLU activation function, before being fed into the final classifying layer for prediction. This method, characterized by a unified analysis and attention-based processing, distinguishes our approach from traditional models and is pivotal to the enhanced predictive capabilities of our framework. Mathematically, this process involves transforming the input vector $\vec{V}_{\text{concat}}$ into an attention vector $\vec{V}_{\text{attention}}$ using the self-attention mechanism, further processing it through a fully connected (FC) layer and a Rectified Linear Unit (ReLU) activation to obtain a refined feature vector $\vec{V}_{fc}$, as outlined below:

\begin{equation}
    \vec{V}_{\text{attention}} = \text{SelfAttention}(\vec{V}_{\text{concat}})
\end{equation}
\begin{equation}
    \vec{V}_{fc} = \text{ReLU(FC}(\vec{V}_{\text{attention}}))
\end{equation}
\begin{equation}
    \vec{z} = \text{Classifier}(\vec{V}_{fc})
\end{equation}

The model leverages these refined features, $\vec{V}_{fc}$, in a classifier to produce logits $\vec{z}$, subsequently processed to predict probabilities for ED Disposition or ED Decompensation tasks. The classifier's output is optimized by minimizing Cross Entropy Loss $L$, ensuring alignment of predicted probabilities $\hat{y}$ with true labels $y_i$. For multi-label tasks like ED Decompensation, each logit $\vec{z}_{i,l}$ undergoes individual sigmoid activation $\sigma$, and the model's training involves minimizing a tailored Cross Entropy Loss that aggregates binary cross-entropy losses across all labels for each observation, capturing the multi-label aspects of the data effectively.

\subsection{Preprocessing \& Model Optimization}
\subsubsection{{Missing data imputation and tokenization}} 

{Tabular EHR are converted into pseudo-notes using predefined templates as described above.} {To represent missing data entries, we incorporate} filler sentences such as \texttt{"The patient did not receive any medications"} in cases where the medication reconciliation (medrecon) modality data is missing. We apply this method to all six modalities before feeding them into our tokenizer derived from the MedBERT model. This model utilizes a subword tokenizer optimized to handle out-of-vocabulary (OOV) words by segmenting words into smaller subword tokens or subword pieces, with a vocabulary size of 28,996 subword tokens. We then split our dataset into training, validation, and testing sets using a fixed seed to ensure the replicability of the results found in this study, thus concluding the data preprocessing phase.

 \begin{figure*}[h!]
   \centering 
   \includegraphics[width=5in]{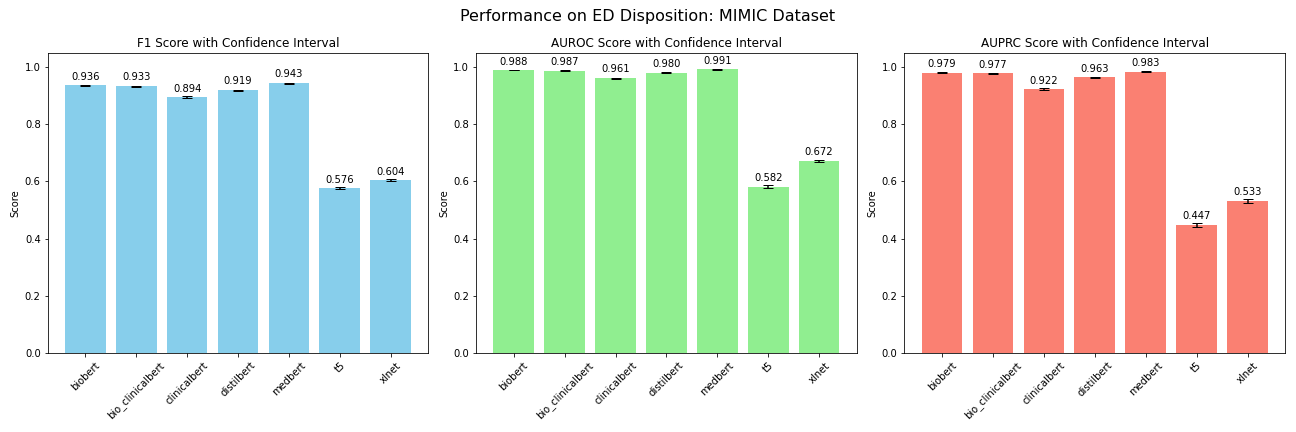} 
    \includegraphics[width=5in]{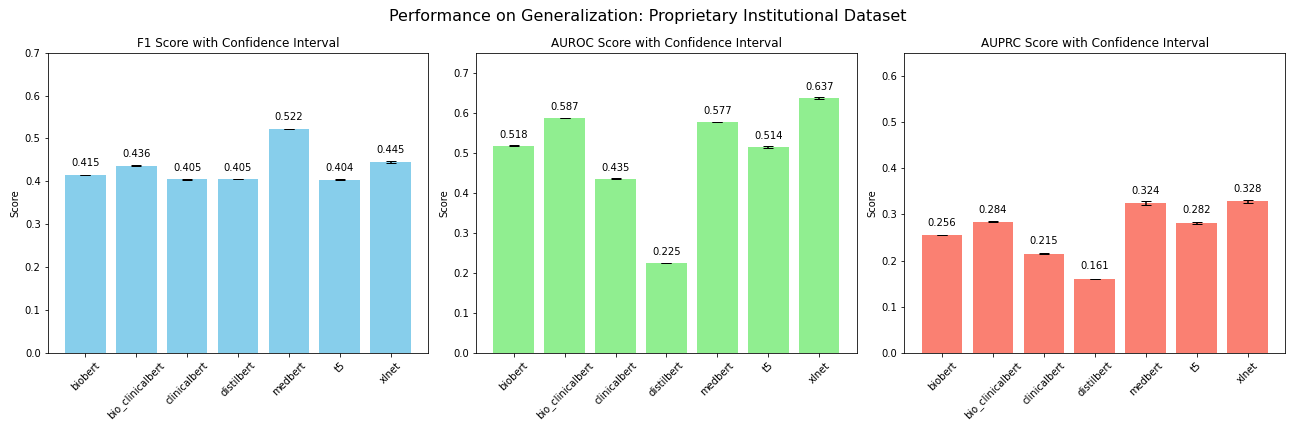} 
   \caption{The benchmark of various foundation models evaluated on ED Disposition. We also explore the model that best generalizes with an additional proprietary institutional dataset for foundation model selection. However, as displayed, all models have trouble with generalization.}
   \label{fm} 
 \end{figure*} 

\subsubsection{Model Optimization}

The models were trained with a batch size of 64, a dropout rate of 0.3, the AdamW optimizer with a learning rate of 5e-5, and a linear learning rate scheduler. For the ED disposition task, we employed Cross-Entropy Loss, and for multilabel decompensation classification, we used Binary Cross-Entropy (BCE) Loss. Training proceeded until a minimum was reached in the validation loss across 10 epochs with early stopping implemented. We tracked F1 scores and loss after each epoch to assess the model's effectiveness. Our computational framework was developed in Python's PyTorch, using large language models available on the HuggingFace \citep{wolf2019huggingface} Platform. For development purposes, we utilized the g4dn.4xlarge EC2 instance from AWS Cloud Services. The model's training and testing were conducted on a single Tesla V100 GPU with 16GB of VRAM to ensure efficient processing.




\section{Results}


\subsection{Model optimization}


\subsubsection{Embedding Model Selection}

We evaluate five different BERT-based models, as well as two other large language models, as potential foundation model backbones for our task, assessing their embedding quality and generalization capabilities. The models under consideration are shown in Appendix Section (\ref{model_des}) in Table \ref{TableModel} with descriptions for each model. We elected to utilize a BERT-based approach due to its bidirectional embeddings, which provide a more complete view of the context around each word. This allows better reasoning about a word's meaning and role for prediction tasks like the ones presented in this work. In contrast, GPT's left-to-right embeddings may miss important contextual cues from future words, limiting their effectiveness for many prediction tasks that require joint understanding of the full context (\cite{ethayarajh2019contextual}; \cite{schomacker2021language}; \cite{topal2021exploring}).

The foundation model backbones are evaluated for their effectiveness and generalization capabilities on our specific ED Disposition reference task, with the goal of selecting the most suitable option for our use case as the multiple embedding model for EHR (MEME). The results are displayed in Figure \ref{fm}. From this analysis we find Medbert \citep{9980157}, which was built on top of \citep{alsentzer2019publicly}, to be the most suitable foundation model.

\subsubsection{Multimodal vs unimodal EHR representation}

\begin{figure}[h!]
   \centering 
   \includegraphics[width=3.5in]{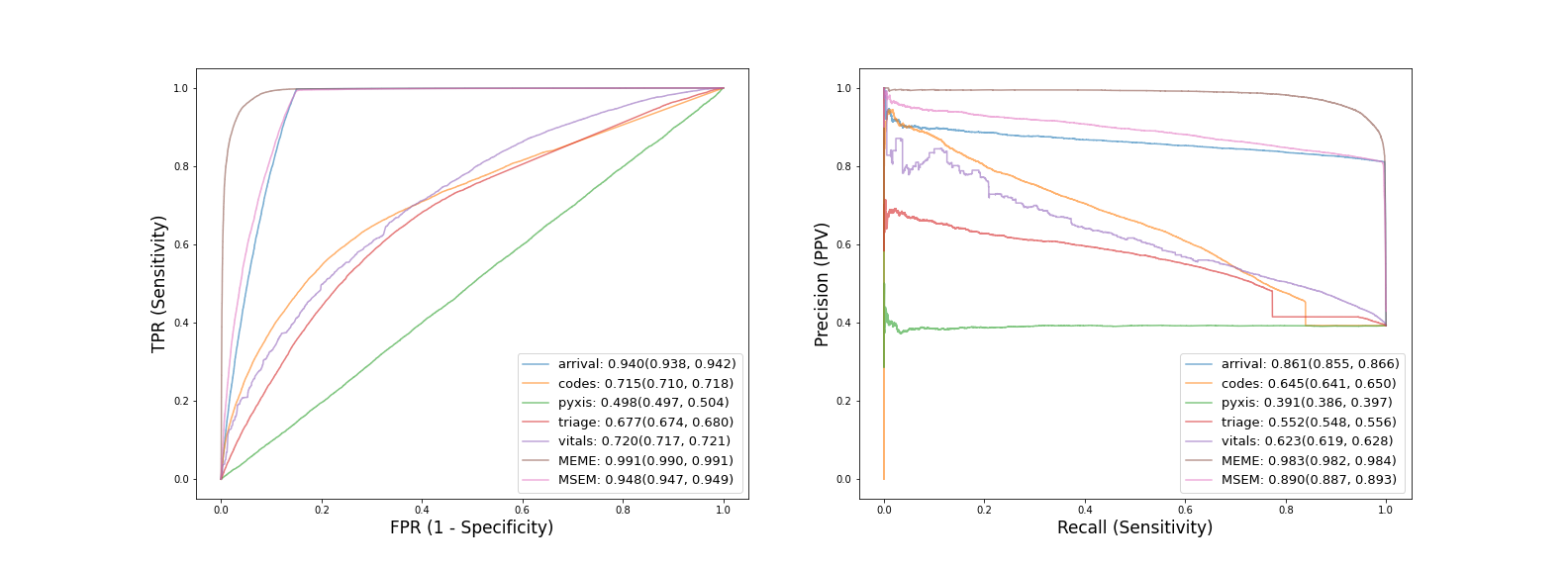} 
   \caption{We performed an ablation study where each modality was used alone to predict the ED Disposition reference task, showcasing the added value of a multimodal data approach.}
   \label{abs} 
 \end{figure} 
\vspace{-0.1cm}
We conduct an ablation study to highlight the benefits of adopting a multimodal data and multiple embedding approach, as illustrated in Figure \ref{abs}. Further results tables can be seen in Appendix Section (\ref{ablation}). This analysis allows us to identify which modalities are most indicative of predicting Emergency Department Disposition based on sets of AUROC and AUPRC results. We observe that no single modality on its own can outperform MEME, as demonstrated through direct comparisons. Additionally, we benchmark MEME against a model designed to encapsulate all multimodal information within a single, heterogeneous embedding. We find that this model's predictive capability is somewhat diminished which is consistent with \citep{rupp_exbehrt_2023} due to truncation and sequence length limitations. This comparison further supports our decision to employ a multiple embedding strategy.

\subsection{MEME vs. reference models}

\begin{table}[H]
\caption{Benchmark study F1. A * denotes that the model operated on tabular data, and a $\dagger$ denotes the model operated on textual (pseudo-notes) data.}
\label{r1}
\begin{adjustbox}{width=\columnwidth}
\begin{small}
\begin{tabular}{l|c|ccc}
\toprule
F1 Benchmark & Dispositon & & Decompensation &\\
Model & ED Disposition & Discharge & ICU & Mortality \\
\midrule
Logistic Regression (Baseline) * &0.799 $\pm$ 0.025&0.549 $\pm$ 0.033& 0.427 $\pm$ 0.036& 0.095 $\pm$ 0.026\\
Random Forest (Baseline) * & 0.826 $\pm$ 0.013 & 0.625 $\pm$ 0.026 & 0.544 $\pm$ 0.048 &  \textbf{0.175 $\pm$ 0.094}  \\
MLP * & 0.841 $\pm$ 0.010 & 0.612 $\pm$ 0.013 & 0.502 $\pm$ 0.019 & 0.097 $\pm$ 0.023 \\
GenHPF \citep{hur2023genhpf} * & --- & --- & ---& --- \\
EHR-Shot\citep{wornow2024ehrshot} *& 0.874 $\pm$ 0.003 & 0.691 $\pm$ 0.008 & 0.560 $\pm$ 0.008 & 0.036 $\pm$ 0.003 \\
MC-BEC\citep{chen2023multimodal} $\dagger$ & 0.912 $\pm$ 0.002& 0.653 $\pm$ 0.006 & 0.545 $\pm$ 0.006 & 0.127 $\pm$ 0.014\\
GPT3.5-turbo $\dagger$ & 0.764 $\pm$ 0.000& --- & --- & ---\\
MSEM $\dagger$& 0.893 $\pm$ 0.003 & 0.622 $\pm$ 0.007 & 0.334 $\pm$ 0.008 & 0.072 $\pm$ 0.014\\
MEME $\dagger$& \textbf{0.943 $\pm$ 0.003} & \textbf{0.698 $\pm$ 0.007} & \textbf{0.572 $\pm$ 0.014} & 0.137 $\pm$ 0.035 \\
\bottomrule
\end{tabular}
\end{small}
\end{adjustbox}
\end{table}
\vspace{-0.6cm}
\begin{table}[H]
\caption{Benchmark study AUROC.}
\label{r2}
\begin{adjustbox}{width=\columnwidth}
\begin{small}
\begin{tabular}{l|c|ccc}
\toprule
AUROC Benchmark & Dispositon & & Decompensation &\\
Model & ED Disposition & Discharge & ICU & Mortality \\
\midrule
Logistic Regression (Baseline) * &0.863 $\pm$ 0.012&0.852 $\pm$ 0.014 & 0.807 $\pm$ 0.017 & 0.768 $\pm$ 0.019\\
Random Forest (Baseline) *& 0.902 $\pm$ 0.010 & \textbf{0.862 $\pm$ 0.014} & \textbf{0.903 $\pm$ 0.016} &   0.847 $\pm$ 0.055\\
MLP *& 0.871 $\pm$ 0.018 & 0.802 $\pm$ 0.011 & 0.767 $\pm$ 0.011 & 0.786 $\pm$ 0.013 \\
GenHPF * & --- & 0.850 $\pm$ 0.000 & ---& \textbf{~0.870 $\pm$ 0.000} \\
EHR-Shot *& 0.790 $\pm$ 0.031 & 0.743 $\pm$ 0.007 & 0.821 $\pm$ 0.18 & 0.827 $\pm$ 0.009\\
MC-BEC $\dagger$& 0.968 $\pm$ 0.002 & 0.708 $\pm$ 0.006 & 0.818 $\pm$ 0.014 & 0.815 $\pm$ 0.006 \\
MSEM$\dagger$& 0.948 $\pm$ 0.002 & 0.552 $\pm$ 0.008 & 0.522 $\pm$ 0.009 & 0.546 $\pm$ 0.023\\
MEME $\dagger$& \textbf{0.991 $\pm$ 0.001} & 0.799 $\pm$ 0.006 & 0.870 $\pm$ 0.015 & \textbf{0.862 $\pm$ 0.006} \\
\bottomrule
\end{tabular}
\end{small}
\end{adjustbox}
\end{table}
\vspace{-0.6cm}
\begin{table}[H]
\caption{Benchmark study AUPRC.}
\label{r3}
\begin{adjustbox}{width=\columnwidth}
\begin{small}
\begin{tabular}{l|c|ccc}
\toprule
AUPRC Benchmark & Dispositon & & Decompensation &\\
Model & ED Disposition & Discharge & ICU & Mortality \\
\midrule
Logistic Regression (Baseline) * &0.874 $\pm$ 0.027&0.628 $\pm$ 0.036& 0.618 $\pm$ 0.034& 0.051 $\pm$ 0.034\\
Random Forest (Baseline) *& 0.902 $\pm$ 0.012 & 0.645 $\pm$ 0.036 & 0.571 $\pm$ 0.047 &  0.102 $\pm$ 0.044  \\
MLP *& 0.866 $\pm$ 0.018 & 0.630 $\pm$ 0.024 & 0.581 $\pm$ 0.026 & 0.077 $\pm$ 0.033 \\
GenHPF * & --- & --- & ---& ---   \\
EHR-Shot *& 0.878 $\pm$ 0.007 & 0.655 $\pm $ 0.012 & 0.655 $\pm$ 0.017 & \textbf{0.246 $\pm$ 0.030} \\
MC-BEC $\dagger$& 0.935 $\pm$ 0.003& 0.657 $\pm$ 0.009 & 0.608 $\pm$ 0.09& 0.174 $\pm$ 0.025 \\
MSEM $\dagger$& 0.890 $\pm$ 0.005 & 0.493 $\pm$ 0.010 & 0.216 $\pm$ 0.009 & 0.037 $\pm$ 0.006\\
MEME $\dagger$& \textbf{0.983 $\pm$ 0.002} & \textbf{0.765 $\pm$ 0.008} & \textbf{0.709 $\pm$ 0.012} & 0.243 $\pm$ 0.034 \\
\bottomrule
\end{tabular}
\end{small}
\end{adjustbox}
\end{table}

In our study, we evaluate the performance of the Multiple Embedding Model for EHR (MEME) against methodologies established by previous research \citep{xie2022benchmarking}, within the domain of Emergency Medicine. This benchmarking study informed comparing various frameworks  previously worked on in the literature as previous state of the art in Emergency Medicine analyses in the case of binary classification. In addition we included several benchmarks from related studies, adopting the methodology of their framework for our tasks.

These comparisons were run within and across datasets. The metrics used for evaluation include the F1 scores, the Area Under the Receiver Operating Characteristic Curve (AUROC), and the Area Under Precision-Recall Curve (AUPRC). To ensure robustness, 95\% confidence intervals were generated for each metric by resampling the test set 1,000 times. Our results—comprised of F1 scores, AUROC scores, and AUPRC scores—are presented in Tables (\ref{r1}, \ref{r2},  \ref{r3}).


\subsubsection{MEME vs traditional ML}
We observed that MEME {generally} surpasses the performance of {traditional techniques operating upon tabular EHR {Tables \ref{r1}, \ref{r2}, \ref{r3}}. We evaluated MEME against a logistic regression, random forest, and neural network model proposed on a previous benchmark \citep{xie2022benchmarking}, operating over tabular EHR prior to pseudonote generation.} Although the Random Forest exhibits a competitive advantage in its AUROC, the imbalanced nature of these tasks makes the benchmarking more nuanced, as it performs inadequately under AUPRC \citep{saito2015precision}.

\subsubsection{MEME vs EHR foundation models}

{EHR-specific foundation models have been recently developed and have shown predictive capabilities across a variety of healthcare applications. We selected the following reference EHR FMs:}

\begin{enumerate}
    \item \textit{GenHPF} \citep{hur2023genhpf}: This specialized model leverages the SimCLR framework for self-supervised learning of representations to predict outcomes. \footnote{\textit{We keep in mind that this method utilizes all the available data from the MIMIC III, MIMIC-IV, and eICU datasets to form predictions while our method only uses that within the MIMIC-IV ED.}}

    \item \textit{EHR-Shot} \citep{wornow2024ehrshot}: Utilizes the CLMBR-T Base Transformer, as proposed by \citep{steinberg2021language}, to generate patient representations from structured electronic health records (EHR) data, sourced from JSON files. These representations are subsequently employed to pursue our classification objectives, mirroring the strategy used in MEME.
    
    \item \textit{MC-BEC} \citep{chen2023multimodal}: Adopts the model architecture outlined by Chen et al., which employs embeddings generated from multiple foundational models (\cite{huang2019clinicalbert}; \cite{yan2022radbert}). These embeddings are then used in a Light Gradient Boosting Model (LGBM) for prediction.
\end{enumerate}
We observed that MEME outperformed  MSEM, MC-BEC, EHR-shot models \footnote{The authors have noted that their CLMBR-t transformer does not support non-OMOP vocabulary contained in the MIMIC-IV database, potentially affecting its performance. We highlight this as a limitation in their EHR-shot model.} in this evaluation. We observe that GenHPF outperforms MEME in the AUROC metric, which can be directly compared with our method. However, it is worth mentioning that their approach incorporates all available EHR data from three databases (MIMIC III, IV, \& eICU) to inform these predictions, and thus test set signal may have leaked into the available model weights. Additionally, it is important to note that AUROC alone may not adequately reflect true performance, particularly in the context of our imbalanced classification objectives.

\subsubsection{MEME vs MSEM}
{Consistent with our prior analysis of multimodal EHR representation, we found that MEME significantly outperformed single-modality embedding (MSEM) across all tasks (Tables \ref{r1}, \ref{r2}, \ref{r3}). A primary rationale for the discrepancy between MEME and MSEM could be attributed to MedBERT's, and more generally BERT architectures', token sequence length, which truncates all input after reaching its 512-sequence limit.

\subsubsection{MEME vs LLM prompting}


{Given the emergent capabilities of generative AI models (e.g. GPT \citep{radford2018improving}, LLaMA-2 \citep{touvron2023llama}, Claude, etc.), we investigated predictive performance of MEME relative to a zero-shot prompting approach.} {We compared the} MEME classifier and a zero-shot GPT-3.5-Turbo API using 100 random samples to predict ED disposition. Although our study's sample size is relatively small, we observed a performance gap displayed in Table \ref{r1}, indicating that training a classifier remains preferable for accurate predictions. An additional plot is attached is shown in the appendix Section \ref{gptee}. Even though GPT-3.5-Turbo exhibited performance beyond a random prediction, our findings underscore the continued importance of our approach.

\subsection{Generalizability of our Method}

A recent critique of healthcare AI applications identified the lack of external validation and therefore potential overfitting to existing publicly available data. Indeed, we observed that while our approach and our reference models displayed strong within-dataset performance using a train-test split, performance was reduced when externally validated across institution (Figure \ref{bar13}).
These results are analogous to the study of \citep{jiang2023health}, who identified that fine-tuning on local hospital systems can improve general performance due to differences caused by demographics, locality, healthcare accessibility and more. 

\begin{figure}[h!]
\begin{center}
\label{bar}
\centerline{\includegraphics[width=2.5in]{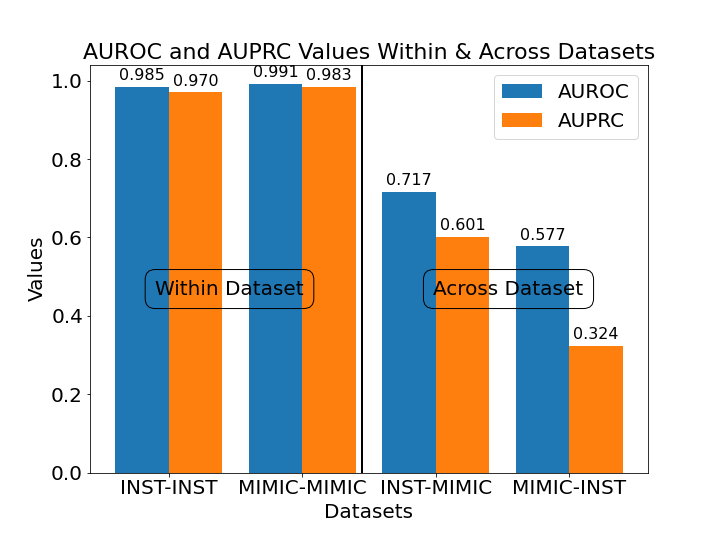}}
\caption{Comparative Analysis of AUROC and AUPRC Values for Intra- and Inter-Dataset Evaluations: This bar chart illustrates the AUROC and AUPRC values of different models, where the models themselves are represented by their training and testing datasets (e.g., MIMIC-MIMIC). \textit{INST stands for institutional dataset}.}
\label{bar13}
\end{center}
\end{figure}
We performed a qualitative error analysis on the top 10 and bottom 10 scoring exemplars in our Institutional dataset to gain insight into this performance gap. Concordant cases were defined as those which the model scored highly and were correctly identified, or scored lowly and correctly not flagged, and Discordant cases were defined as vice versa. No systematic patterns or patient attributes were immediately apparent from investigating the decision making of this model in these groups, necessitating future study. 

However, we observed a distribution shift between datasets in terms of EHR features. {We compared the unique tabular values across the two EHR databases and observed only moderate overlap between values and concepts} (Figure \ref{venn}). For a more detailed perspective see appendices (Section \ref{gener3}). This might reflect systematic differences in patient populations or hospital protocols. 

\begin{figure}[h!]
\vskip 0.2in
\begin{center}
\label{bar}
\centerline{\includegraphics[width=2.5in]{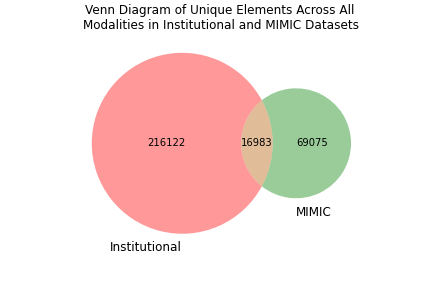}}
\caption{Venn Diagram of Unique Features across all modalities respective of both EHR Systems. Note the large differences in features indicating a distribution shift.}
\label{venn}
\end{center}
\vskip -0.2in
\end{figure}

\section{Discussion \& Conclusion}

In this paper we introduce {a Multimodal Embedding Model for EHR (MEME), a representation framework for EHR. MEME is built upon two core concepts: 1) The text serialization of tabular EHR into pseudo-notes, which provides an interface with language foundation models; and 2) the multimodal framing of EHR to reflect the heterogeneity of the underlying data.} 
{We demonstrate that this combination outperforms traditional ML techniques, EHR-specific foundation models, and prompting-based approaches across decision support tasks in the Emergency Department.} 

{In addition to the performance advantages demonstrated by our experiments, MEME has several qualitative benefits in terms of portability and extendibility. EHR-specific models, such as BEHRT \citep{li_behrt_2020}, CHIRoN \citep{hill2023chiron}, EHR-shot \citep{wornow2024ehrshot}, etc., rely on data standards and transparent harmonization procedures to ensure interoperability, and their general applicability is severely hampered by the fact that these standards are still being developed and evaluated (e.g., MEDS\_ETL vs OMOP). It is also unclear how these models should be extended to accommodate new and updated medical concepts \citep{arnrich2024medical}. By contrast, the natural language approach is extendible to any data that can be text-serialized, which is more easily adopted by institutions, and can more gracefully handle changes in coding standards, all the while leveraging general reasoning capabilities and increasing medical domain knowledge captured by LLMs. While all of our findings were demonstrated within the Emergency Department setting, we hypothesize that they should generalize into other common decision support scenarios.}

\paragraph{Limitations}

One crucial limitation of the model is the poorly performing external validity caused by the heterogenous nature of two EHR datasets. 
We saw in our results that it is likely that protocols around patient care vary across site and over time such that the same patient would experience different outcomes depending on when and where they experienced care. Therefore we conclude that current models developed on the publicly available MIMIC-IV ED dataset appear to be insufficient for ensuring true generalization across diverse healthcare systems.

Another limitation of this work is our inability to release our private institutional data, due to privacy restrictions and university policy. This highlights the significance of independent benchmarks, and underscores the necessity of external validation, including benchmark datasets and tasks such as MC-BEC \citep{chen2023multimodal}.
\paragraph{Future Works}
Our Pseudo-notes approach can be extended beyond classification tasks. These applications encompass, but are not limited to, Retrieval Augmented Generation (RAG), recommendation systems, summarization, and other clinical tasks associated with textual modalities. Additionally, another promising direction could involve pre-training a Large Language Model from scratch utilizing pseudo-notes sourced from multiple institutions. This approach aims to build a pre-trained model fundamentally based on structured Electronic Health Records, potentially enhancing its relevance and efficacy in healthcare contexts.

\subsection{Impact Statement}

The goal of this work is to advance the field of Machine Learning in Healthcare, and thus presents a novel potential interface between modern NLP (LLMs) and clinical data. However, this work does not directly address issues involving performance and bias of all forms within LLMs and thus potentially introduces these issues into healthcare applications. This, coupled with our limited inter-site performance, highlight an urgent need for inter-site validation and further study before these models can leave the laboratory. 

\paragraph{Code and Data} Code, Data, and MIMIC-IV ED based model weights on HuggingFace will be made available in the camera ready version.



\bibliography{example_paper}
\bibliographystyle{icml2024}

\newpage
\appendix
\onecolumn
\section{Appenidx}

\subsection{Strobe Diagrams of Our Data}
\label{strober}

We include Strobe Diagrams of our two datasets to give the breakdown numbers of our data pictured if Figure \ref{strobe}.

 \begin{figure}[b!]
   \centering 
   \includegraphics[width=4.5in]{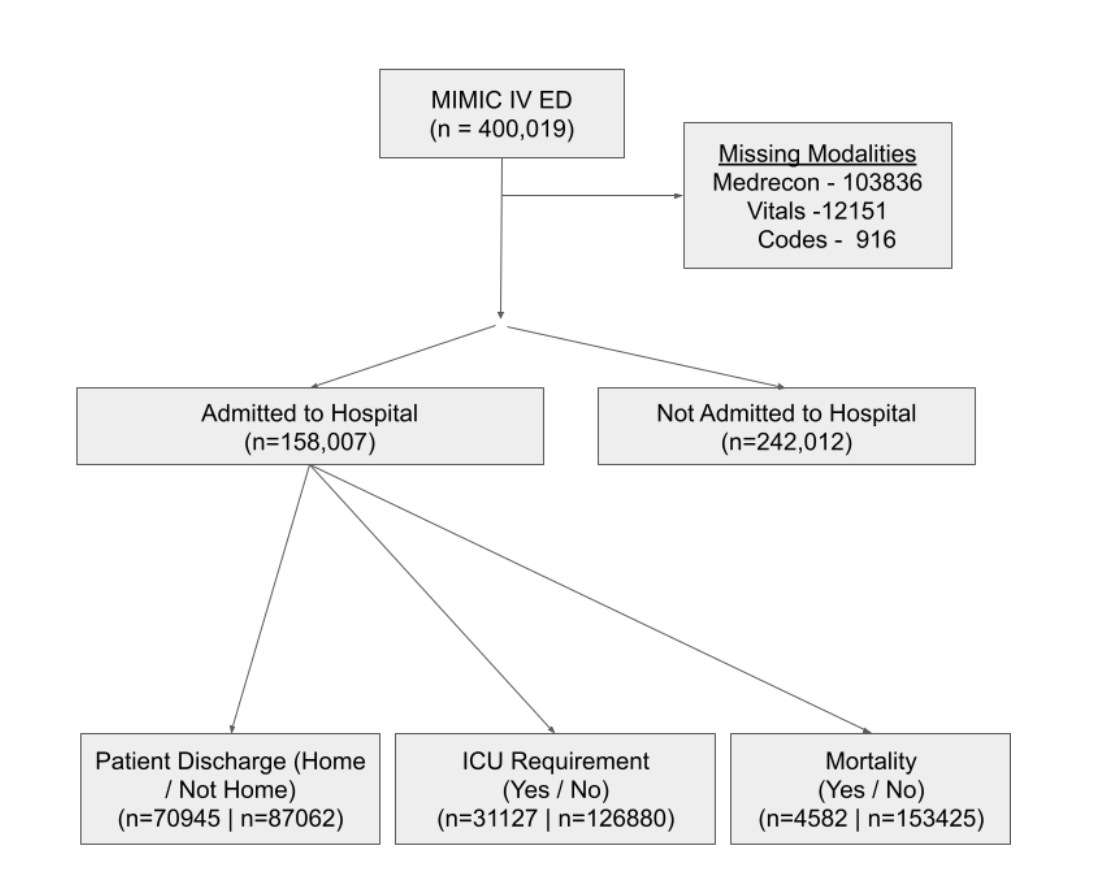} 
   \includegraphics[width=4.5in]{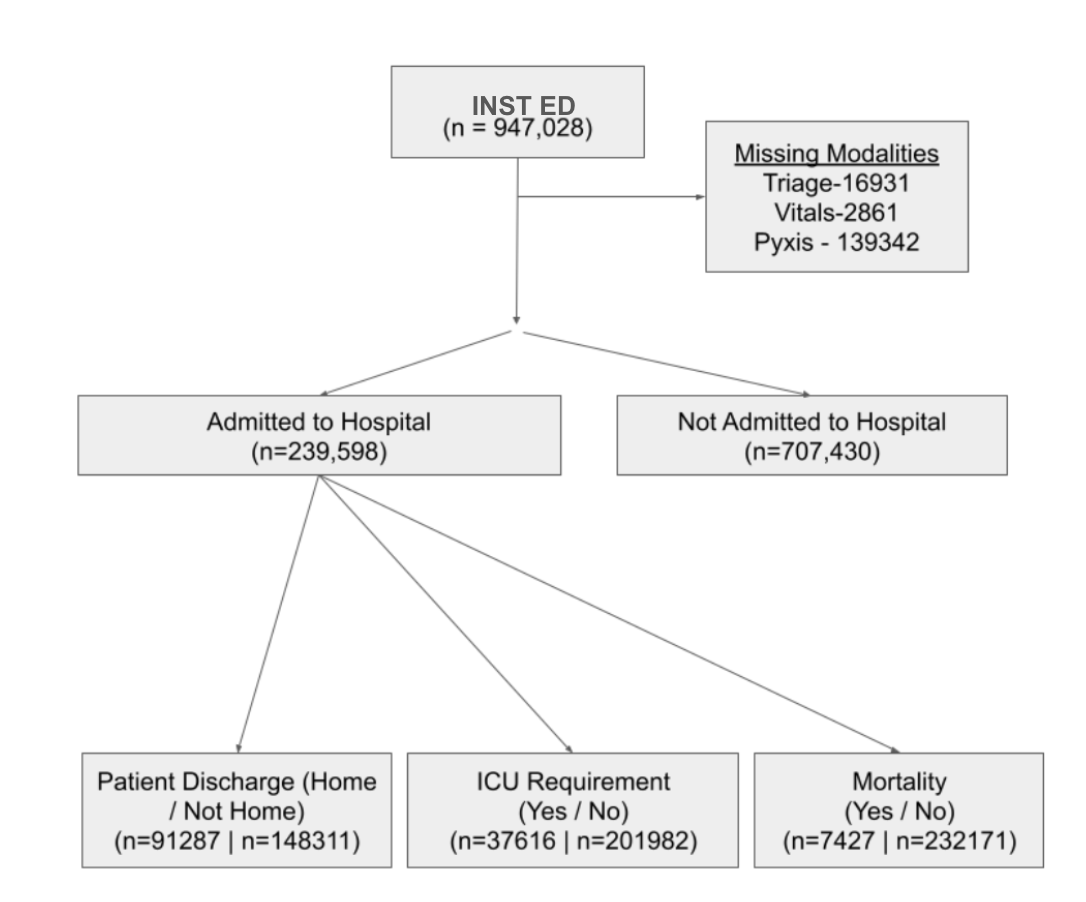} 
   \caption{Strobe Diagrams for both MIMIC-IV ED and our Institutional ED Dataset}
   \label{strobe} 
 \end{figure} 
\newpage
\subsection{Example Pseudonotes}
\label{exnotes}

In Section \ref{notes}, we referred to the appendix to provide a verbose exploration of our pseudo-notes generation process, offering more detailed and illustrative examples. Building upon the foundation laid out earlier, we utilize color coordination to clearly delineate the  distinctions between information sourced directly from our raw EHR data and content that was dynamically generated through our script.

\begin{multicols}{2}

\textit{Arrival Information}\\
\fbox{\begin{minipage}{15em}
Patient \textcolor{red}{10000032}, a \textcolor{red}{52} year old \textcolor{red}{white female}, arrived via \textcolor{red}{ambulance} at \textcolor{red}{2180-05-06 19:17:00}. The patient's marital status is \textcolor{red}{widowed}. The patient's insurance is \textcolor{red}{other}. The patient's language is \textcolor{red}{english}.
\end{minipage}}\\

\textit{Emergency Department Disposition}\\
\fbox{\begin{minipage}{15em}
The ED disposition was \textcolor{red}{admitted} at \textcolor{red}{2180-05-06 23:30:00}. The patient died on \textcolor{red}{2180-09-09}.
\end{minipage}}\\

\textit{Triage}\\
\fbox{\begin{minipage}{15em}
At triage: temperature was \textcolor{red}{98.4}, pulse was \textcolor{red}{70}, respirations was \textcolor{red}{16}, o2 saturation was \textcolor{red}{97}, systolic blood pressure was \textcolor{red}{106}, diastolic blood pressure was \textcolor{red}{63}, pain was \textcolor{red}{0}, chief complaint was \textcolor{red}{abd pain}, \textcolor{red}{abdominal distention}. Acuity score was \textcolor{red}{3}.
\end{minipage}}\\

\textit{Medrecon}\\ 
\fbox{\begin{minipage}{15em}
The patient was previously taking the following medications: \textcolor{red}{albuterol sulfate}, \textcolor{red}{asthma/copd therapy - beta 2-adrenergic agents}, \textcolor{red}{inhaled}, \textcolor{red}{short acting}. \textcolor{red}{peg 3350-electrolytes}, \textcolor{red}{laxative - saline/osmotic mixtures}. \textcolor{red}{nicotine}, \textcolor{red}{smoking deterrents - nicotine-type}. \textcolor{red}{spironolactone [aldactone]}, \textcolor{red}{aldosterone receptor antagonists}. \textcolor{red}{emtricitabine-tenofovir [truvada]}, \textcolor{red}{antiretroviral - nucleoside and nucleotide analog rtis combinations}. \textcolor{red}{raltegravir [isentress]}, \textcolor{red}{antiretroviral - hiv-1 integrase strand transfer inhibitors}. \textcolor{red}{spironolactone [aldactone]}, \textcolor{red}{diuretic - aldosterone receptor antagonist}, \textcolor{red}{non-selective}. \textcolor{red}{furosemide, diuretic - loop}. \textcolor{red}{ipratropium bromide [atrovent hfa]}, \textcolor{red}{asthma/copd - anticholinergic agents}, \textcolor{red}{inhaled short acting}. \textcolor{red}{ergocalciferol (vitamin d2)}, \textcolor{red}{vitamins - d derivatives}.
\end{minipage}} 

\textit{Patient Vitals}\\
\fbox{\begin{minipage}{15em}
The patient had the following vitals: At \textcolor{red}{2180-05-06 23:04:00}, temperature was \textcolor{red}{97.7}, pulse was \textcolor{red}{79}, respirations was \textcolor{red}{16}, o2 saturation was \textcolor{red}{98}, systolic blood pressure was \textcolor{red}{107}, diastolic blood pressure was \textcolor{red}{60}, pain was \textcolor{red}{0}.
\end{minipage}} \\

\textit{Pyxis}\\
\fbox{\begin{minipage}{15em}
The patient received the following medications: At \textcolor{red}{2180-08-05 22:29:00}, \textcolor{red}{morphine} were administered. At \textcolor{red}{2180-08-05 22:55:00}, \textcolor{red}{donnatol (elixir), aluminum-magnesium hydrox.-simet, aluminum-magnesium hydrox.-simet, ondansetron, ondansetron} were administered. 
\end{minipage}}\\

\textit{Diagnostic Codes}\\
\fbox{\begin{minipage}{15em}
The patient received the following diagnostic codes: ICD-9 code: \textcolor{red}{[78959], other ascites}. ICD-9 code: \textcolor{red}{[07070], unspecified viral hepatitis c without hepatic coma}. ICD-9 code: \textcolor{red}{[5715], cirrhosis of liver nos}. ICD-9 code: \textcolor{red}{[v08], asymptomatic hiv infection}. 
\end{minipage}}

\end{multicols}

\subsection{Foundation Model Descriptions for Model Selection Experiment}
\label{model_des}

A Table of the foundation models that we used to select the MEME model along with their descriptions are found in Table \ref{TableModel}.

\begin{table}[H]
\centering
\caption{The foundation models evaluated for becoming the backbone for MEME.}
\begin{tabular}{p{6cm}p{8cm}}
\toprule
\textbf{Model} & \textbf{Description} \\
\midrule
\textbf{BioBERT} \citep{lee2020biobert} & A BERT variant pretrained on large biomedical corpora, enabling it to better capture domain-specific linguistic patterns and outperform general BERT on biomedical text mining tasks like named entity recognition, relation extraction, and question answering.\\
\midrule
\textbf{Bio\_ClinicalBERT} \citep{alsentzer2019publicly} & Initialized with BioBERT and further pretrained on clinical notes from the MIMIC III database (~880M words), this model is designed for tasks involving electronic health records from ICU patients.\\
\midrule
\textbf{ClinicalBERT} \citep{huang2019clinicalbert} & Pretrained on a large clinical corpus of 1.2B words spanning diverse diseases and fine-tuned on over 3 million electronic health records, this model is well-suited for clinical NLP tasks due to its understanding of medical terminology and health record data.\\
\midrule
\textbf{DistilBERT} \citep{sanh2019distilbert} & A lightweight and faster version of BERT, designed for efficient inference on resource-constrained devices while retaining most of BERT's performance on various NLP tasks.\\
\midrule
\textbf{MedBERT} \citep{9980157} & A model initialized with Bio\_ClinicalBERT and further pretrained on biomedical datasets like N2C2, BioNLP, and CRAFT, tailored for biomedical named entity recognition tasks involving diseases, drugs, genes, and other healthcare concepts.\\
\midrule
\textbf{T5} \citep{2020t5} & Pretrained on a massive text corpus using a text-to-text denoising objective, this encoder-decoder transformer model can be flexibly applied to various NLP tasks like translation, summarization, and question answering by framing them as text-to-text problems.\\
\midrule
\textbf{XLNet} \citep{yang2019xlnet} & A generalized autoregressive pretraining method that combines the advantages of autoregressive language modeling and the permutation-based approaches used in BERT. It captures bidirectional contexts by maximizing the expected likelihood over all permutations of the input sequence factorization order.\\
\bottomrule
\label{TableModel}
\end{tabular}
\end{table}

\newpage
\subsection{Ablation Study Table}
\label{ablation}

We display the raw performance metrics from our ablation study of the various modalities contributing to the multiple tasks proposed in our work on the MIMIC IV ED dataset. The complete set of results is displayed in Tables \ref{r4}, \ref{r5}, \ref{r6}}. We note that no modality alone beat out the MEME model which took on a multimodal approach. 

\begin{table}[h!]
\centering
\caption{F1 Scores for MIMIC-IV ED Dataset Ablation Study Across Different Tasks.}
\label{r4}
\begin{adjustbox}{width=5in}
\begin{small}
\begin{sc}
\begin{tabular}{l|c|ccc}
\toprule
F1 Benchmark & Dispositon & & Decompensation &\\
Modality & ED Disposition & Discharge & ICU & Mortality \\
\midrule
Arrival & 0.895 $\pm$ 0.003 & 0.533 $\pm$ 0.008 & 0.041 $\pm$ 0.009 & 0.063 $\pm$ 0.005 \\
Codes & 0.613 $\pm$ 0.003 & 0.572 $\pm$ 0.008 & 0.054 $\pm$ 0.009 & 0.055 $\pm$ 0.004 \\
Medrecon & 0.625 $\pm$ 0.005 & 0.525 $\pm$ 0.009 & 0.029 $\pm$ 0.005 & 0.054 $\pm$ 0.004 \\
Pyxis & 0.564 $\pm$ 0.004 & 0.478 $\pm$ 0.009 & 0.167 $\pm$ 0.015 & 0.060 $\pm$ 0.005 \\
Triage & 0.596 $\pm$ 0.005 & 0.403 $\pm$ 0.010 & 0.367 $\pm$ 0.015 & 0.055 $\pm$ 0.004 \\
Vitals & 0.619 $\pm$ 0.005 & 0.366 $\pm$ 0.009 & 0.160 $\pm$ 0.013 & 0.067 $\pm$ 0.006 \\
MEME & \textbf{0.943 $\pm$ 0.003} & \textbf{0.698 $\pm$ 0.007} & \textbf{0.572 $\pm$ 0.014} & \textbf{0.137 $\pm$ 0.035} \\
\bottomrule
\end{tabular}
\end{sc}
\end{small}
\end{adjustbox}
\end{table}

\begin{table}[H]
\caption{AUROC Scores for MIMIC-IV ED Dataset Ablation Study Across Different Tasks.
}
\centering
\label{r5}
\begin{adjustbox}{width=5in}
\begin{small}
\begin{tabular}{l|c|ccc}
\toprule
AUROC Benchmark & Dispositon & & Decompensation &\\
Modality & ED Disposition & Discharge & ICU & Mortality \\
\midrule
Arrival & 0.940 $\pm$ 0.002 & 0.689 $\pm$ 0.007 & 0.682 $\pm$ 0.009 & 0.709 $\pm$ 0.019 \\
Codes & 0.715 $\pm$ 0.005 & 0.658 $\pm$ 0.007 & 0.714 $\pm$ 0.008 & 0.709 $\pm$ 0.020 \\
Medrecon & 0.709 $\pm$ 0.004 & 0.644 $\pm$ 0.007 & 0.622 $\pm$ 0.009 & 0.668 $\pm$ 0.021 \\
Pyxis & 0.498 $\pm$ 0.005 & 0.616 $\pm$ 0.008 & 0.704 $\pm$ 0.009 & 0.705 $\pm$ 0.024 \\
Triage & 0.677 $\pm$ 0.004 & 0.647 $\pm$ 0.007 & 0.758 $\pm$ 0.008 & 0.736 $\pm$ 0.020 \\
Vitals & 0.720 $\pm$ 0.004 & 0.598 $\pm$ 0.008 & 0.733 $\pm$ 0.008 & 0.740 $\pm$ 0.021 \\
MEME & \textbf{0.991 $\pm$ 0.001} & 0.\textbf{799 $\pm$ 0.006} & \textbf{0.870 $\pm$ 0.015} & \textbf{0.862 $\pm$ 0.006} \\
\bottomrule
\end{tabular}
\end{small}
\end{adjustbox}
\end{table}

\begin{table}[H]
\centering
\caption{AUPRC Scores for MIMIC-IV ED Dataset Ablation Study Across Different Tasks.}
\label{r6}
\begin{adjustbox}{width=5in}
\begin{small}
\begin{tabular}{l|c|ccc}
\toprule
AUPRC Benchmark & Dispositon & & Decompensation &\\
Model & ED Disposition & Discharge & ICU & Mortality \\
\midrule
Arrival & 0.861 $\pm$ 0.005 & 0.625 $\pm$ 0.010 & 0.376 $\pm$ 0.014 & 0.076 $\pm$ 0.013 \\
Codes & 0.645 $\pm$ 0.007 & 0.595 $\pm$ 0.011 & 0.410 $\pm$ 0.014 & 0.061 $\pm$ 0.008 \\
Medrecon & 0.589 $\pm$ 0.007 & 0.579 $\pm$ 0.010 & 0.286 $\pm$ 0.010 & 0.049 $\pm$ 0.006 \\
Pyxis & 0.391 $\pm$ 0.005 & 0.561 $\pm$ 0.011 & 0.484 $\pm$ 0.015 & 0.112 $\pm$ 0.022 \\
Triage & 0.552 $\pm$ 0.008 & 0.579 $\pm$ 0.011 & 0.504 $\pm$ 0.015 & 0.100 $\pm$ 0.017 \\
Vitals & 0.623 $\pm$ 0.007 & 0.544 $\pm$ 0.010 & 0.430 $\pm$ 0.014 & 0.087 $\pm$ 0.014 \\
MEME & \textbf{0.983 $\pm$ 0.002} & \textbf{0.765 $\pm$ 0.008} & \textbf{0.709 $\pm$ 0.012} & \textbf{0.243 $\pm$ 0.034} \\

\bottomrule
\end{tabular}
\end{small}
\end{adjustbox}
\end{table}
\newpage
\subsection{GPT3.5-Turbo Prompt}
\label{gptee}

\begin{lstlisting}[language=Python, caption=Basic Example of How prompted the gpt-3.5-turbo model to generate predictions.]
from openai import OpenAI
client = OpenAI()

completion = client.chat.completions.create(
  model="gpt-3.5-turbo",
  messages=[
    {"role": "system", "content": "You are a medical officer, skilled in determining whether a patient should be admitted to the Emergency Room or not."},
    {"role": "user", "content": "I have 10 patient samples here. I need you to predict whether each patient should be admitted to the emergency room or not. Give you prediction in the list format ([`1,0,0,1,1,1,0,1,0,1']) and predict 1 if they should be admitted and 0 if not:\n\n'Patient 10000032, a 52 year old white female, arrived via ambulance at 2180-05-06 19:17:00. The patient's marital status is widowed. The patient's insurance is other. The patient's language is english.The patient received the following diagnostic codes: ICD-9 code: [5728], oth sequela, chr liv dis. ICD-9 code: [78959], other ascites. ICD-9 code: [07070], unspecified viral hepatitis c without hepatic coma. ICD-9 code: [v08]...'

    ...
    [OTHER_PSEUDONOTES]
    ...
    "}
  ]
)

print(completion.choices[0].message)

>>> Here are the predictions: {"predictions": [1, 0, 1, 1, 1, 1, 1, 1, 1, 1]}
\end{lstlisting}

We conducted a benchmarking study comparing whether training a model was necessary with the emergence of generative AI like (GPT, LLaMA, etc.). Here, we present a bar graph of the performance of MEME relative to prompting based approach. We find that training a classifier remains preferable.

\begin{figure}[h!]
    \centering
    \includegraphics[width=3in]{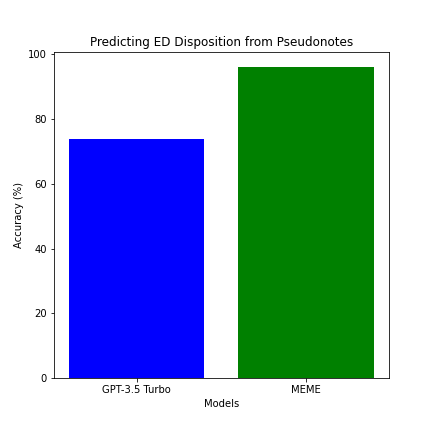}
    \caption{Comparison of Classification Accuracies: Training a Classifier (MEME) versus Utilizing a Generative Model (GPT-3.5-Turbo) for Prediction}
    \label{gpt-meme}
\end{figure}

\subsection{Why did Generalization Fail?}
\label{gener3}


To investigate the failure of our generalization experiment from MIMIC to our institutional database, we conducted a thorough examination of the data input into our pseudo-notes method. We analyzed the data from both Electronic Health Record (EHR) sources by plotting distributions via box plots, and Venn diagrams to elucidate the disparities between the datasets.

\subsubsection*{Arrival Information}

\begin{figure}[h!]
   \centering 
   \includegraphics[width=4.5in]{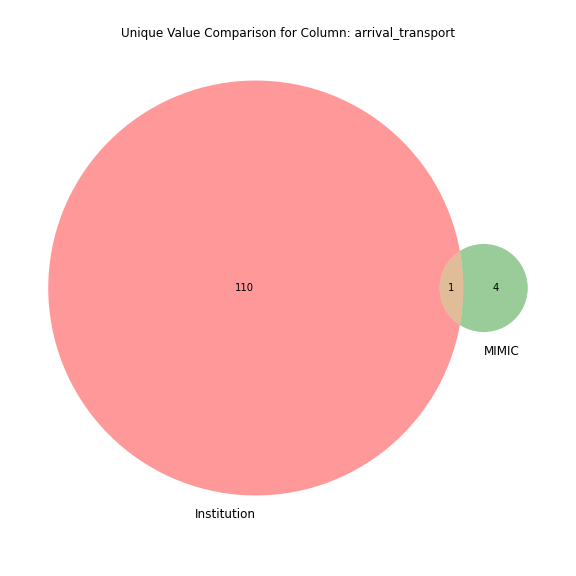} 
   \includegraphics[width=4.5in]{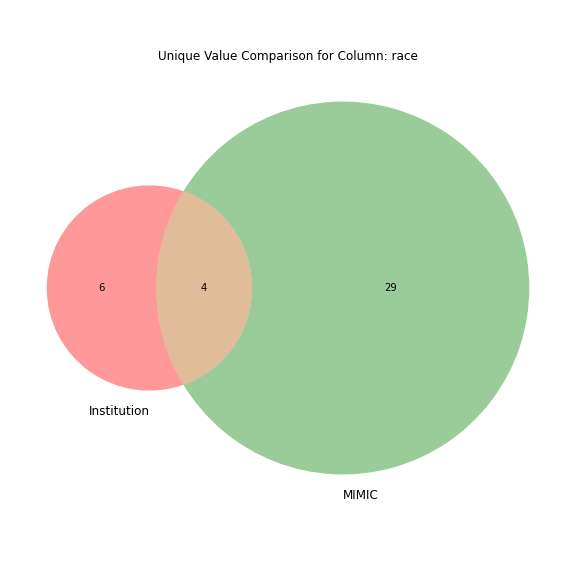} 
   \caption{The Venn diagram illustrates the overlap between unique elements identified in our Institutional Data and MIMIC-IV ED \textbf{Arrival} Information Data.}
   \label{app1} 
 \end{figure}

All results related to \textit{arrival} information are presented in Figure \ref{app1}. We observe notable differences within these extensive categorical classes for arrival transport and race labels. Although these differences are not considered substantial, highlighting these disparities remains critical.

 \subsubsection*{Diagnoses}
\begin{figure}[h!]
   \centering 
   \includegraphics[width=4.5in]{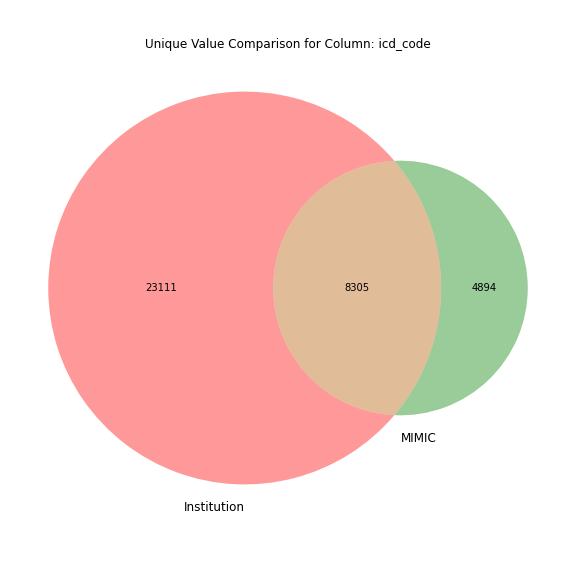} 
   \includegraphics[width=4.5in]{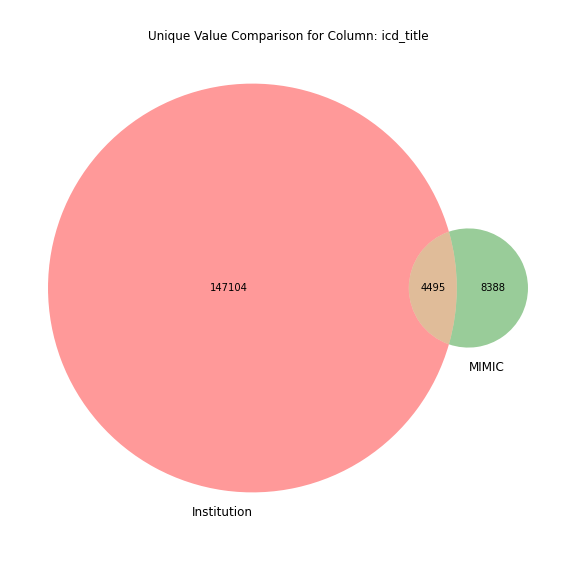} 
   \caption{The Venn diagram illustrates the overlap between unique elements identified in our Institutional Data and MIMIC-IV ED \textbf{Diagnoses} Data.}
   \label{app2} 
 \end{figure}

All results pertaining to \textit{diagnoses }information are presented in Figure \ref{app2}. We observe notable differences within the ICD Diagnostic codes. These differences are critical and may contribute to failed generalization.
 
 \subsubsection*{Pyxis}
\begin{figure}[h!]
   \centering 
   \includegraphics[width=5in]{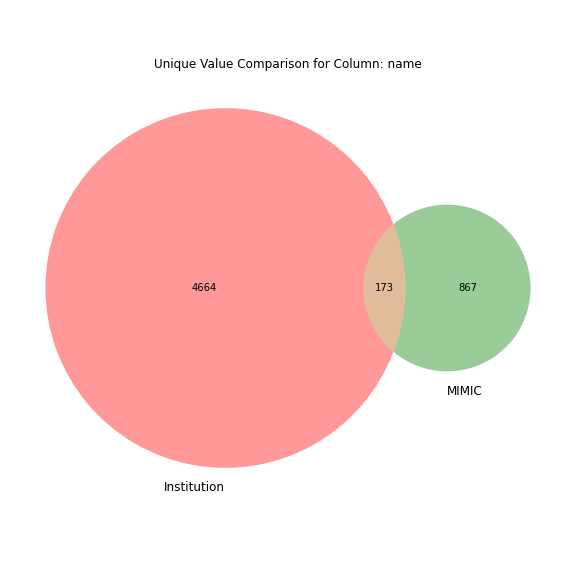} 
   \caption{The Venn diagram illustrates the overlap between unique elements identified in our Institutional Data and MIMIC-IV ED \textbf{Pyxis} Data.}
   \label{app3} 
 \end{figure}

 All results pertaining to \textit{pyxis} information are presented in Figure \ref{app3}. We observe notable differences within the drug names dispensed during their ED Visit. These differences are critical and may contribute to failed generalization.

 \subsubsection*{Triage}
\begin{figure}[h!]
   \centering 
   \includegraphics[width=2.5in]{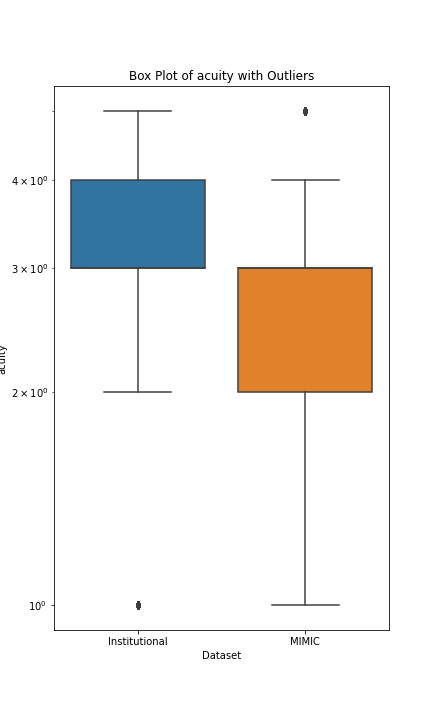} 
   \includegraphics[width=2.5in]{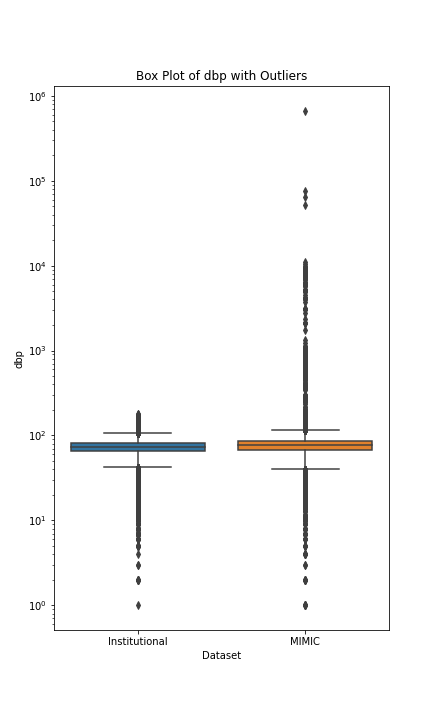} 
   \includegraphics[width=2.5in]{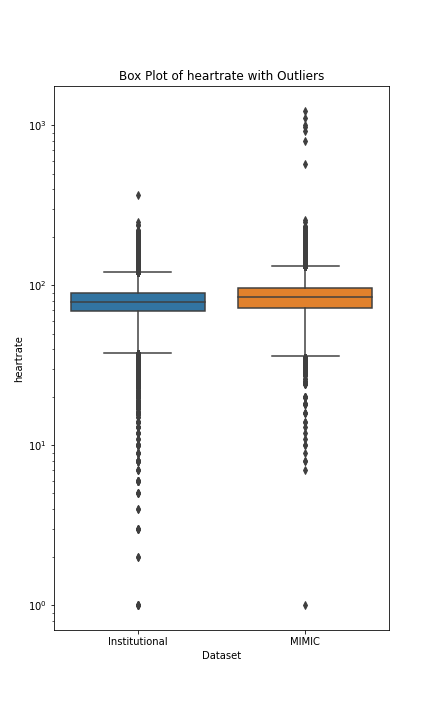} 
   \includegraphics[width=2.5in]{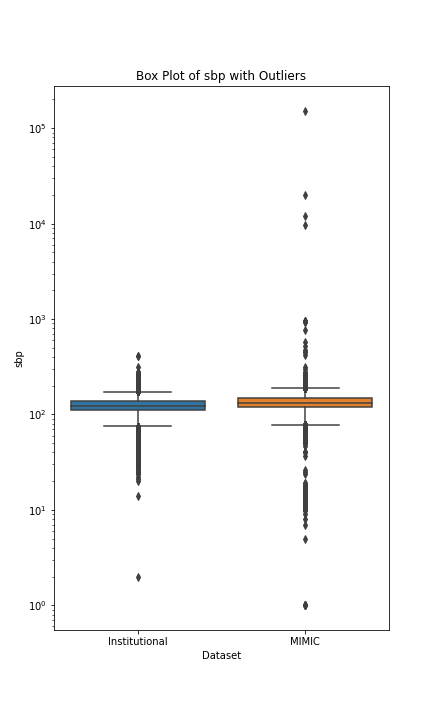} 
   \caption{The box plot depicts the distribution of values within Institutional and MIMIC-IV ED \textbf{Triage Information} Datasets, thereby facilitating a comparative analysis of data variability and central tendency across the datasets.}
   \label{app4} 
 \end{figure}

\begin{figure}[h!]
   \centering 
   \includegraphics[width=5in]{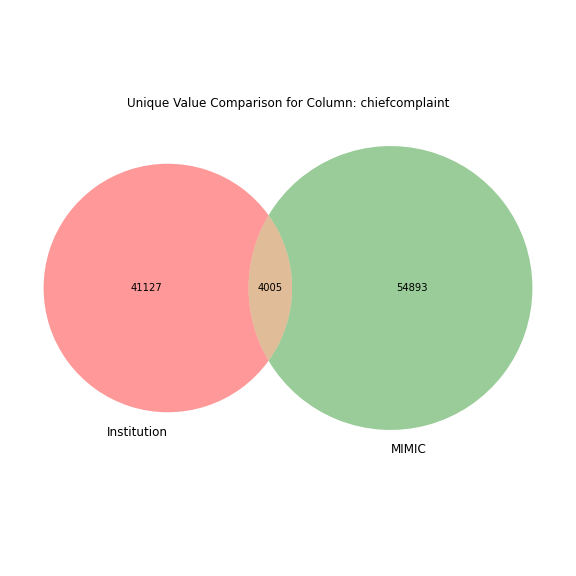} 
   \caption{The Venn diagram illustrates the overlap between unique elements identified in our Institutional Data and MIMIC-IV ED \textbf{Chief Complaints} Data coming from \textbf{Triage} Information.}
   \label{app5} 
 \end{figure}

All results related to \textit{pyxis} information are presented in Figures \ref{app4} and \ref{app5}. We observe notable differences in the chief complaints, which were anticipated to vary. However, we emphasize that numerical values within EHRs appeared as expected and exhibited similar distributions with marginal outliers. These differences are critical and may contribute to failed generalization.

 \subsubsection*{Vitals}
\begin{figure}[h!]
   \centering 
   \includegraphics[width=2.5in]{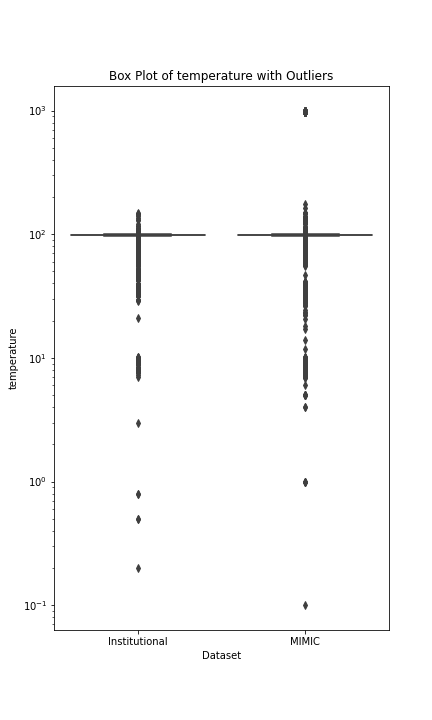} 
   \includegraphics[width=2.5in]{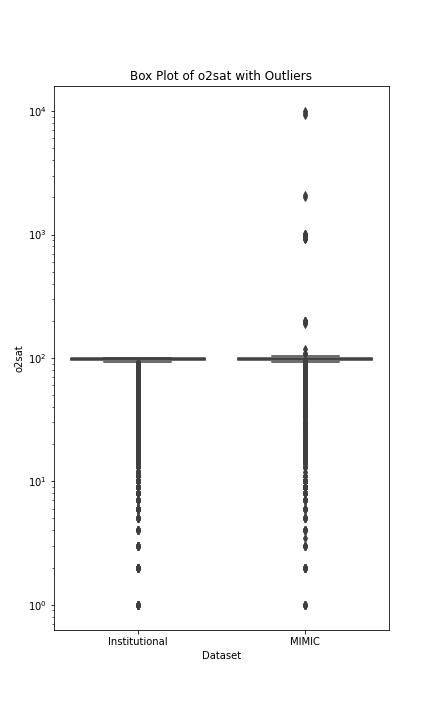} 
   \includegraphics[width=2.5in]{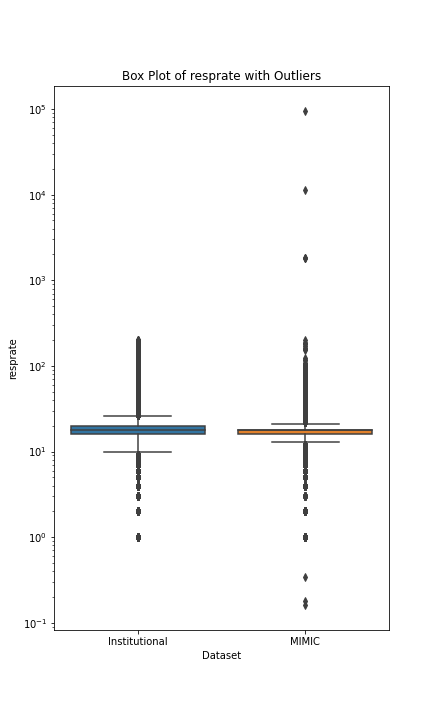} 
   \caption{The box plot depicts the distribution of values within Institutional and MIMIC-IV ED \textbf{Vitals Information} Datasets, thereby facilitating a comparative analysis of data variability and central tendency across the datasets.}
   \label{app6} 
 \end{figure}

All results related to \textit{vitals} information are presented in Figures \ref{app6}. We see again that numerical data appears to follow similar distributions despite a few outliers. We do not think the numerical aspects of the EHR led to failed generalization.

\subsection*{Remarks based on Analysis}

Our analysis reveals that EHR systems exhibit significant variability especially in the case of free text and categorical data types. This diversity is evident in the unique chief complaints, diagnostic annotations, and other unique medical cases of different hospital institutions, which contribute to the challenges in achieving generalization. These insights underscore the complexities inherent in working across heterogeneous EHR environments and advise to use this method with some local fine-tuning of hospitals as highlighted in \citep{jiang2023health}.
\end{document}